%% file: main.tex
\documentclass[conference]{IEEEtran}
\IEEEoverridecommandlockouts

\PassOptionsToPackage{numbers,sort&compress}{natbib}
\usepackage{natbib}

\usepackage[utf8]{inputenc}
\usepackage[T1]{fontenc}
\usepackage{amsmath,amssymb,amsfonts}
\usepackage{graphicx}
\usepackage{booktabs}
\usepackage{multirow}
\usepackage{xcolor}
\usepackage{url}
\usepackage{microtype}
\usepackage[hypcap=true]{caption}
\usepackage{hyperref}

\usepackage{algorithm}
\usepackage{algpseudocode}
\usepackage{siunitx}

\usepackage{tikz}
\usetikzlibrary{positioning, shapes.geometric, arrows.meta, fit}
\usepackage{pgfplots}
\pgfplotsset{compat=1.18}
\usepgfplotslibrary{groupplots}

\usepackage{placeins}

\usepackage{flushend}

\usepackage{titlesec}
\titleformat{\section}
  {\normalfont\large\bfseries}{\thesection}{0.8em}{}
\titleformat{\subsection}
  {\normalfont\normalsize\bfseries}{\thesubsection}{0.8em}{}
\titleformat{\subsubsection}
  {\normalfont\normalsize\bfseries\itshape}{\thesubsubsection}{0.8em}{}
\renewcommand{\thesection}{\arabic{section}}
\renewcommand{\thesubsection}{\arabic{section}.\arabic{subsection}}
\renewcommand{\thesubsubsection}{\arabic{section}.\arabic{subsection}.\arabic{subsubsection}}

\title{Code Isn't Memory: A Structural Codebase Index Inside a Coding Agent%
\thanks{Code and data: \url{https://github.com/TransformerOptimus/supercoder-eval}}}

\author{%
\IEEEauthorblockN{\textbf{Ishaan Bhola}}
\IEEEauthorblockA{\textit{SuperAGI Research}}
\and
\IEEEauthorblockN{\textbf{Adithyan Krishnan}}
\IEEEauthorblockA{\textit{SuperAGI Research}}
\and
\IEEEauthorblockN{\textbf{Sravanth Kurmala}}
\IEEEauthorblockA{\textit{SuperAGI Research}}
\and
\IEEEauthorblockN{\textbf{Mukunda NS}}
\IEEEauthorblockA{\textit{SuperAGI Research}}
}

\begin{document}

\maketitle

\input{sections/00_abstract}
\input{sections/01_introduction}
\input{sections/02_related_work}
\input{sections/03_system}
\input{sections/04_experimental_design}

\input{sections/05_integrity}
\input{sections/06_results}
\input{sections/07_discussion}

\bibliographystyle{unsrtnat}
\bibliography{references}

\end{document}

%% file: sections/00_abstract.tex
%
%

\begin{abstract}
Coding agents now interleave LLMs with retrieval over the working repository, and
retrieval implementations vary widely across deployed harnesses. Inside a fixed
coding-agent harness on a fixed model, does adding a structural codebase index actually
change cost or resolve? We ran three arms (the harness with the index, the same harness
without it, and an agentic-grep comparator) on SWE-PolyBench Verified and SWE-bench Pro
with Claude Opus~4.7 \citep{claudeopus47} held fixed throughout, across three seeds, inside a leak-audited
per-task sandbox. The within-harness ablation produces a large localization
gain and a statistically separated resolve gain, with no cost penalty per cell and lower
cost per solve. The cross-harness check shows that the index does not regress against an
agentic-grep baseline on resolve or localization, again at no cost penalty. We release
the per-cell exclusion ledger, the leak-audit script, the localization extractor, and the
results database. The deployment question for a structural codebase index is thus not
whether it is too expensive to run (across seeds, the index lands at a lower \$/solved
than agentic grep) but whether the workload includes multi-file changes where structural
ranking pays off.
\end{abstract}

\vspace{0.3em}

%% file: sections/01_introduction.tex
%
%

\section{Introduction}
\label{sec:intro}

Coding agents now interleave LLMs with retrieval over the working repository, and retrieval
implementations vary widely across deployed harnesses. The implementations span a
spectrum: agentic grep over the working copy, file-dependency repo maps, semantic and
graph search, and structural codebase indices built once per repository
(\S\ref{sec:related}). Inside a fixed coding-agent harness on a fixed model, does adding a
structural codebase index actually change cost or resolve? We answer the question for
open-source harnesses with the model held fixed (Claude Opus~4.7
\citep{claudeopus47}, \S\ref{sec:design-model}); closed-source harnesses (Claude Code, Cursor, Windsurf) are out
of scope by design. The experimental design isolates the index causally by toggling it on
and off inside one harness while everything else stays identical, and cross-checks the
result against an agentic-grep comparator (\S\ref{sec:design-arms}).

The field has not had a clean answer because controlled measurements are scarce. Most
prior work either compares whole harnesses, where retrieval is confounded with prompt,
tool surface, and control loop, or evaluates retrieval components in isolation against
acc@$k$, without the downstream agentic loop that turns ranking into a fix. The
question whether ``grep is all you need'' has been asked recently in the agentic-search
literature for memory-style document retrieval \citep{grepallyouneed2026}, with grep
favored over vector retrieval; we ask the code-task counterpart, where the candidate
beyond grep is not a vector index but a structural codebase index (semantic + lexical +
call-graph). A structural codebase index is also expensive to build and operate, so if
the resolve gain is small and the cost premium is large, the index does not pay for
itself in a deployment. The integrity bar for benchmark evaluation has risen in
parallel: recent audits documented solution leakage in issue text
\citep{swebenchplus2024}, memorization of in-benchmark repositories
\citep{swebenchillusion2025}, and substantial score inflation from formal issue text
relative to realistic user phrasing \citep{savingswebench2025}, so any positive result
needs to survive a leak audit before it counts.

Three arms ran against the same SWE-PolyBench Verified and SWE-bench Pro public instances
(91 instances; Go, Java, Python) with Claude Opus~4.7 fixed throughout, across three seeds:
SC-ON (the SuperCoder \citep{supercoder2024} harness with the index on), SC-OFF (the same
harness with the two engine tools removed, every other component identical), and OpenCode
\citep{opencode2025} (an agentic-grep comparator). Every cell ran
inside a hardened per-task sandbox with a fail-closed git scrub and a post-run leak audit
(\S\ref{sec:integrity}). On the causal within-harness ablation
(\S\ref{sec:results-ablation}), the index moves View~B acc@5 from 44.3\% to 84.5\% across
seeds (paired Wilcoxon $p<0.0001$) and resolve from 41.9\% to 50.4\% (paired Wilcoxon
$p=0.003$), and yields lower cost per solve with a statistically null per-cell cost
difference. On the cross-harness validity check
(\S\ref{sec:results-crossharness}), SC-ON matches or modestly favors OpenCode on resolve
(50.4\% vs.\ 45.3\% mean, paired Wilcoxon $p=0.087$) and on View~B acc@5
(84.5\% vs.\ 75.3\% mean, paired Wilcoxon $p=0.080$) at no cost penalty. The
structural codebase index does not duplicate behavior that competent agentic grep already
reaches; at minimum, it does not regress the agent.

We report the first leak-audited, model-controlled, causal ablation of a shipped
structural codebase index inside a coding-agent harness, paired with a cross-harness
validity check against an agentic-grep comparator. The result reframes the deployment
question from ``is a structural codebase index too expensive to run alongside agentic
grep'' (on these benchmarks, the answer is no: \$2.30 mean across seeds against
OpenCode's \$2.92, favorable on \$/solved) to ``does the workload include multi-file
changes where structural ranking pays off'' (\S\ref{sec:discussion}). Released alongside
the paper are the per-cell exclusion ledger (\S\ref{sec:integrity-ledger}), the leak-audit
script, the dual-view localization extractor, and the full results database, so every
number in \S\ref{sec:results} is reproducible from the released artifacts.

%% file: sections/02_related_work.tex
%
%

\section{Related Work}
\label{sec:related}

\textbf{Coding-agent harnesses.} SWE-agent \citep{sweagent2024} introduced the
agent-computer-interface framing on top of a single-LLM control loop; OpenHands
\citep{openhands2025}, formerly OpenDevin, generalized the platform with sandboxed
execution and multi-agent coordination; Aider \citep{aider} drives file-level edits over a
local git repository with a dependency-ranked repo map; AutoCodeRover
\citep{autocoderover2024} pairs LLM reasoning with AST-aware code search and
spectrum-based fault localization; OpenCode \citep{opencode2025} is the model-agnostic
open-source TUI agent we use as the cross-harness comparator
(\S\ref{sec:system-opencode}). SuperCoder, the harness this paper studies
(\S\ref{sec:system-harness}), shares the parallel-tool-dispatch loop posture with these
systems but ships a structural codebase index as a first-class tool, which prior
measured harnesses do not. We exclude SWE-agent from the comparator set because its
agent-computer-interface pipeline was used in SWE-bench's construction, creating
circularity for an evaluation on SWE-bench-family tasks. Closed-source harnesses (Claude
Code, Cursor, Windsurf) are out of scope by design; this study fixes the model
(\S\ref{sec:design-model}) and varies the open-source harness configuration around it.

\textbf{Retrieval approaches for code agents.} Four approach types appear in the recent
literature. \emph{Agentic grep and read} drives OpenCode \citep{opencode2025} and similar
terminal-loop agents that call ripgrep over the working copy; there is no structural
codebase index. Sen et al.\ \citep{grepallyouneed2026} contrast grep with vector retrieval
inside agentic loops on the LongMemEval memory-retrieval benchmark, with grep favored;
their setting is non-code and the alternative they ablate against is dense retrieval, not
a structural codebase index, but the question framing (does grep suffice inside an
agentic harness?) is the closest prior to ours. \emph{Repository-level retrieval and
planning} approaches predate the agent-harness wave: RepoCoder \citep{repocoder2023}
iteratively retrieves over the whole repository for code completion, and CodePlan
\citep{codeplan2024} stages multi-file edits as a planned sequence of repository-wide
operations --- neither is built as an agent loop. \emph{File-dependency repo maps},
exemplified by Aider's PageRank-ranked repo map
\citep{aider}, surface candidate files by import and reference structure but do not index
symbol-level semantics. \emph{Semantic and graph search} combines code-chunk embeddings
with typed repository graphs: LocAgent \citep{locagent2025} equips an LLM agent with
graph-search tools over a heterogeneous code graph, RepoGraph \citep{repograph2024} plugs
a repository-wide code graph into SWE-agent and AutoCodeRover, and the Code Graph Model
line \citep{cgm2025} integrates the graph directly into an LLM's attention via an
adapter; Agentless \citep{agentless2024} reaches comparable SWE-bench-Lite scores with a
non-agentic three-phase localization-and-repair pipeline. \emph{Structural codebase
indices} have been adopted in commercial coding-agent stacks; this paper provides the
first leak-audited, model-controlled causal ablation of one inside an open-source harness
(\S\ref{sec:results-ablation}).

\textbf{Localization metrics for code agents.} LocAgent \citep{locagent2025} and
Agentless \citep{agentless2024} report file-level acc@$k$ as the primary localization
metric. The field has been moving toward stage-decomposed trajectory metrics: TRAJEVAL
\citep{trajeval2026} decomposes agent trajectories into search, read, and edit phases
with per-stage precision and recall, and SWE-Explore \citep{sweexplore2026} isolates
repository exploration as a sub-task with coverage and ranking metrics against
trajectory-derived ground truth. Our View~B
(\S\ref{sec:design-localization}, \S\ref{sec:results-crossharness}) sits in the same
field move: we strip engine-result paths from the surfaced set so that an SC-ON acc@$k$
counts the same kind of agent-targeted surface as an OpenCode acc@$k$. This paper does
not propose a new localization metric; it adopts the field-trend rule and applies it
uniformly across arms.

\textbf{SWE-bench family and benchmark integrity.} SWE-bench \citep{swebench2024}
introduced the canonical 2{,}294-issue Python benchmark; SWE-bench Verified
\citep{swebenchverified2024} released a 500-issue human-screened subset. Two recent
benchmark-expansion lines extend coverage: SWE-PolyBench \citep{swepolybench2025} is a
multi-language SWE-bench-style benchmark with a Verified subset, and SWE-bench Pro
\citep{swebenchpro2025} provides a harder long-horizon set; we use the Verified subset of
SWE-PolyBench and the public subset of SWE-bench Pro
(\S\ref{sec:design-benchmarks}). Three integrity audits motivate our hardening protocol:
SWE-bench+ \citep{swebenchplus2024} documented solution leakage in issue text and
weak-test pass-throughs in the original SWE-bench; the SWE-bench Illusion paper
\citep{swebenchillusion2025} showed that strong scores partly reflect memorization of
in-benchmark repositories; and Garg et al.\ \citep{savingswebench2025} mutate the formal
GitHub-issue specification into realistic user-style queries derived from chat-agent
telemetry, and report capability overestimation above 50\% for some models on
public benchmarks --- a phrasing-ecological-validity threat that detect-and-exclude
hardening (this paper) does not address. The per-cell scrub, fail-closed gate, and S1
reviewer pass in \S\ref{sec:integrity} sit in this audit tradition; the public exclusion
ledger (\S\ref{sec:integrity-ledger}) extends it by releasing per-cell evidence for every
dropped cell.

%% file: sections/03_system.tex
%
%
%

\section{System}
\label{sec:system}

This section describes the studied subject: the SuperCoder coding-agent harness,
the context engine that the ON/OFF arms ablate, and the OpenCode comparator that
the cross-harness arm runs.

\subsection{SuperCoder harness}
\label{sec:system-harness}

SuperCoder \citep{supercoder2024} is a coding agent built around a single-LLM control loop. The
shipped binary supports three modes (Ask, Plan, Coding); the evaluation runs in Coding
mode, the only mode that may write files or execute shell commands, so all
mechanics described below refer to the Coding-mode loop. A provider gateway sits
in front of the LLM client and captures token-level cost per turn uniformly across
arms (\S\ref{sec:design-sandbox}).

Each turn assembles a prompt (system instructions, tool schema, message history)
and issues a single LLM call. If the model emits tool calls, the harness
dispatches them, awaits results, and appends them to the message history; if it
emits text with no tool calls, the loop terminates. The reasoning-and-acting
posture follows the ReAct \citep{react2023} pattern, and the tool-call interface
follows the function-calling line introduced by Toolformer \citep{toolformer2023}.
Multiple tool calls in a single response are executed in parallel. If the rolling token count exceeds a
threshold, the harness compacts the message history by summarizing older turns;
the compaction step is disclosed for reproducibility but is not load-bearing for
the ablation. The loop terminates when (a) the model emits no tool calls,
(b) a configured per-cell turn budget is exhausted, or (c) the 30-minute
per-cell wall-clock cap (\S\ref{sec:design-sandbox}) elapses.

The agent calls a fixed tool set: \texttt{read}, \texttt{write}, \texttt{edit},
\texttt{bash}, \texttt{git}, \texttt{grep}, and \texttt{glob}, plus task-management
tools (\texttt{todo\_write}, \texttt{apply\_patch}). All of these are identical
across SC-ON and SC-OFF. The two context-engine tools, \texttt{codebase\_search}
and \texttt{codebase\_graph}, are available only in SC-ON; SC-OFF removes those
two tools from the schema and changes nothing else
(\S\ref{sec:system-engine}).

\subsection{Context engine}
\label{sec:system-engine}

The context engine is a separate service that the agent calls through two tools.
It maintains a per-repository index that is built once on first contact and
updated incrementally on subsequent runs via Merkle-tree diffs over the working
copy, so a source edit invalidates and re-indexes only the affected chunks
rather than the whole repository. Each cell in this study starts from a fresh
sandbox, so every arm exercises the build path; the incremental-update path is
part of the engine but not load-bearing in the run. The index has three
components: a \textbf{vector index} of code-chunk embeddings for semantic
similarity, a \textbf{graph index} of definitions and call edges for structural
reachability, and a \textbf{lexical (BM25) index} of identifiers and tokens for
exact-match recall. Index construction begins with \texttt{tree-sitter} parsing
per source file; definitions, references, and call edges are extracted from the
resulting AST and chunked for embedding.

\begin{figure}[!tbp]
  \centering
  \resizebox{\linewidth}{!}{%
  \begin{tikzpicture}[
      x=1.3cm, y=1.15cm,
      every node/.style={font=\small},
      box/.style={draw, rounded corners=3pt, align=center,
                  inner sep=6pt, minimum height=13mm,
                  text width=36mm, font=\small},
      idx/.style={box, fill=black!5},
      tool/.style={box, fill=black!12},
      arr/.style={-Latex, thick, line width=0.7pt},
      phaselbl/.style={font=\small\bfseries\itshape, fill=white, inner sep=3pt},
    ]
    \node[phaselbl] at (0, 5.0) {Indexing (build once, then Merkle-diff updates)};

    \node[box] (repo)    at (0, 4.0) {\textbf{Repository}\\\scriptsize source files};
    \node[box] (parse)   at (0, 2.7) {\textbf{\texttt{tree-sitter} parse}\\\scriptsize AST per source file};
    \node[box] (extract) at (0, 1.4) {\textbf{Symbol + call-graph extraction}\\\scriptsize definitions, identifiers, call edges};
    \node[box] (chunk)   at (0, 0.1) {\textbf{Chunking + embedding}\\\scriptsize code chunks $\to$ vectors};

    \draw[arr] (repo)    -- (parse);
    \draw[arr] (parse)   -- (extract);
    \draw[arr] (extract) -- (chunk);

    \node[idx] (bm25)   at (-3.4,-1.6) {\textbf{Lexical (BM25)}\\\scriptsize identifier + token matches};
    \node[idx] (graph)  at ( 0,  -1.6) {\textbf{Call-graph index}\\\scriptsize caller / callee edges};
    \node[idx] (vector) at ( 3.4,-1.6) {\textbf{Vector index}\\\scriptsize semantic similarity};

    \draw[arr] (chunk.south) -- (bm25.north);
    \draw[arr] (chunk.south) -- (graph.north);
    \draw[arr] (chunk.south) -- (vector.north);

    \draw[dashed, gray!70] (-5.0, -2.8) -- (5.0, -2.8);
    \node[phaselbl] at (0, -3.1) {Retrieval (per agent call)};

    \node[tool] (tools)   at (-3.4, -4.6) {\textbf{\texttt{codebase\_search} /}\\\textbf{\texttt{codebase\_graph}}\\\scriptsize agent issues query};
    \node[box]  (hybrid)  at ( 0,   -4.6) {\textbf{Hybrid retrieval}\\\scriptsize fuse, rerank, dedup};
    \node[box]  (results) at ( 3.4, -4.6) {\textbf{Ranked results}\\\scriptsize paths + snippets + scores};

    \draw[arr] (tools)  -- (hybrid);
    \draw[arr] (hybrid) -- (results);

    \draw[arr] (bm25.south)   -- (hybrid.north west);
    \draw[arr] (graph.south)  -- (hybrid.north);
    \draw[arr] (vector.south) -- (hybrid.north east);
  \end{tikzpicture}%
  }
  \caption{Context-engine pipeline. The upper block runs once per repository on
    first contact and re-runs incrementally on subsequent contacts via
    Merkle-tree diffs over the working copy: \texttt{tree-sitter} produces an
    AST per source file, an extractor walks the ASTs to collect definitions,
    identifiers, and call edges, and code chunks are embedded into vectors;
    the result is three indices populated in parallel. The lower block runs on
    every agent call:
    \texttt{codebase\_search} or \texttt{codebase\_graph} dispatches a query to
    hybrid retrieval, which fuses hits across the three indices and returns a
    ranked result list to the agent. Per-tool input and result schemas are
    described in \S\ref{sec:system-engine}.}
  \label{fig:engine-pipeline}
\end{figure}
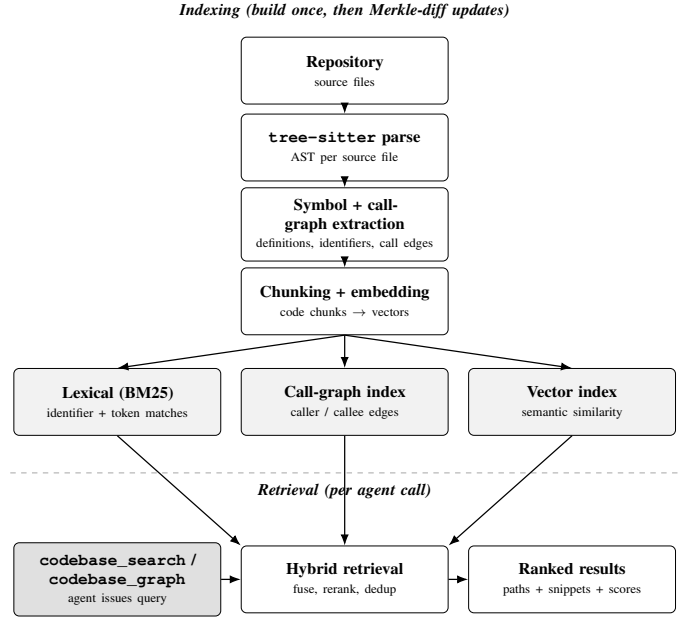

Figure~\ref{fig:engine-pipeline} sketches the indexing pipeline and the retrieval
path. The components are named at the level the public eval repo can support
(\S\ref{sec:integrity-ledger}); the backend service that hosts the three indices
is internal and not part of the released artifact.

\textbf{Agent-facing tools.} \texttt{codebase\_search} takes a natural-language query plus a retrieval strategy (vector, lexical, graph, or hybrid) and returns a ranked list of code chunks, each carrying file path, snippet, relevance score, and the index that produced it; a local overlay drops paths the agent has deleted and flags stale ones. \texttt{codebase\_graph} takes a symbol and traverses the call-graph index, returning callers and callees grouped by direction, each carrying the defining file path and the distance from the query node.

\textbf{ON versus OFF concretely.} SC-ON exposes both
\texttt{codebase\_search} and \texttt{codebase\_graph} to the agent alongside
the file I/O, shell, and search tools listed in \S\ref{sec:system-harness};
SC-OFF removes exactly those two tools from the schema and leaves everything
else identical (the rest of the tool set, the model, the prompt template, the
sandbox, the scorer; \S\ref{sec:design-arms}). The toggle is the headless
runner's command-line flag for the engine endpoint: present invokes SC-ON,
absent invokes SC-OFF. The ablation surface is therefore ``with versus without
the two engine tools,'' and the resolve, localization, and cost consequences of
that surface live in \S\ref{sec:results-ablation}.

\subsection{Comparator: OpenCode}
\label{sec:system-opencode}

OpenCode \citep{opencode2025} is an open-source coding agent: a single-LLM control loop with parallel tool dispatch and a fixed tool set built around \texttt{rg} (ripgrep), \texttt{read}, \texttt{glob}, and \texttt{bash}; no structural codebase index, no embedding-based search, no precomputed call-graph. In our evaluation, OpenCode runs in the same per-task container as the two SuperCoder arms, with Claude Opus~4.7 and the same 30-minute wall-clock cap (\S\ref{sec:design-sandbox}). The headline cross-harness comparison is \S\ref{sec:results-crossharness}.
\FloatBarrier

%% file: sections/04_experimental_design.tex
%
%

\section{Experimental Design}
\label{sec:design}

This section defines the arms, the model, the benchmarks, the run scope, the sandbox and
cost-capture infrastructure, the metrics, the localization-view extraction rule, the
statistical methods, and the narrowed pilot. \S\ref{sec:results} consumes these
definitions verbatim.

\subsection{Arms}
\label{sec:design-arms}

We compare three arms on the same instance set: \textbf{SC-ON} (SuperCoder harness with
the context engine's tools available), \textbf{SC-OFF} (same harness, same prompts, with
\texttt{codebase\_search} and \texttt{codebase\_graph} removed from the toolset), and
\textbf{OpenCode} (an independent open-source harness whose retrieval is built around
\texttt{ripgrep} and file reads, with no structural index). The only thing
that changes between SC-ON and SC-OFF is the engine toolset, which gives the ablation its
causal reading. The cross-harness comparison against OpenCode tests whether SC-ON's
behavior is reproducible by an alternative open-source harness running the same model.

Across all three arms we hold the model (Claude Opus~4.7), the per-task sandbox image, the scorer, and the 30-minute wall-clock cap fixed; no turn or dollar cap is enforced. Each arm runs three seeds. SC-ON exposes the two engine tools (\texttt{codebase\_search}, \texttt{codebase\_graph}); SC-OFF removes exactly those two from the toolset and leaves everything else identical; OpenCode is an independent harness with its own tool surface (\S\ref{sec:system-opencode}).

\subsection{Model}
\label{sec:design-model}

All three arms run Claude Opus 4.7 \citep{claudeopus47} (\texttt{claude-opus-4-7}) across three seeds. Fixing the
model removes capability as a moving part, so any cross-harness difference has to come
from the harness or its retrieval, not from a stronger backbone. Single-model scope is a
limitation we acknowledge in \S\ref{sec:integrity} and \S\ref{sec:discussion}; the
control trade is intentional.

\subsection{Benchmarks}
\label{sec:design-benchmarks}

The instance set draws from two public benchmarks: SWE-PolyBench Verified
\citep{swepolybench2025} contributes the multi-language coverage (Go, Java, Python), and
SWE-bench-Pro \citep{swebenchpro2025} contributes longer-context Python tasks. Both
descend from the SWE-bench family \citep{swebenchverified2024}. We do not use SWE-Agent
or its trajectories as a comparator because of its role in benchmark construction; the
open-source harness comparator is OpenCode \citep{opencode2025}.

\subsection{Run scope}
\label{sec:design-scope}

The study runs on 91 instances across three languages: 34 Go, 20 Java, 37
Python. JavaScript and TypeScript are not covered (limitation logged in
\S\ref{sec:integrity}). The run uses three seeds, pass@1 per seed; statistics are
reported as the mean of seed means with across-seed standard deviation as a variance
estimate, and seed-variance context follows \citep{bjarnason2026randomness}.

\paragraph{Paired-$n$ denominators.}
\S\ref{sec:results} uses three paired-$n$ values, and each one carries a specific
meaning. The \emph{triple-intersection} set is the subset of instances on which all
three arms produced a legit cell ($n=75$); this set is used for descriptive cross-arm
context where all three rows of a table need to refer to the same instances. Pairwise
significance tests use the \emph{pairwise} denominator instead, because dropping
instances that are legit in two arms simply because the third arm failed wastes paired
signal. The pairwise denominators are 80 (SC-ON vs.\ SC-OFF) and 78 (SC-ON vs.\
OpenCode). Every paired test in \S\ref{sec:results} cites the $n$ that applies to it.

\subsection{Sandbox and cost capture}
\label{sec:design-sandbox}

Each cell ran inside a per-task isolated container with a uniform image across arms, on
an internal sandbox backend. The backend's configs and image spec are not part of the
public release. A unified provider gateway captured token-level cost on every LLM call,
which means \$/cell and \$/solved are computed against the same accounting for all three
arms. Per-cell \texttt{cost\_usd}, \texttt{total\_cost\_usd}, \texttt{tokens\_total}, and
\texttt{wall\_clock\_secs} are released in the public DB. The 30-minute wall-clock
cap fired on two cells in the released set (one SC-OFF and one OpenCode); SC-ON's
longest legit cell ran 19 minutes. The released reproducibility kit lives under
\texttt{data/}; per-cell patches, per-trace JSON, prompts, and sandbox image references
are held back because they include licensed repository source and internal harness
configuration. Resolve is the public DB's \texttt{resolved} column, sourced from the
upstream benchmark scorers; \$/cell, \$/solved, turns, tokens, and wall-clock are
released per-cell.

\subsection{Metrics}
\label{sec:design-metrics}

The primary outcome is \textbf{resolve}; supporting outcomes are localization, effort,
and cost. Table~\ref{tab:metrics} states the formal definitions. Localization is
reported under View~B by default (\S\ref{sec:design-localization}); effort and cost
metrics are per-cell means on legit cells from the unified provider gateway
(\S\ref{sec:design-sandbox}).

\begin{table}[!tbp]
  \centering
  \caption{Metric definitions. Resolve is the public DB's \texttt{resolved} column,
    sourced from the upstream benchmark scorers (F2P denotes fail-to-pass tests; P2P
    denotes pass-to-pass tests). Localization is computed under View~B
    (\S\ref{sec:design-localization}); effort and cost are per-cell means on legit cells.}
  \label{tab:metrics}
  \input{tables/tab-metrics}
\end{table}

\FloatBarrier
\subsection{Localization views}
\label{sec:design-localization}

\begin{algorithm}[!tbp]
  \footnotesize
  \caption{Localization extraction. The trace is a sequence of message events; each
  event carries a tool name, a \emph{role} ($\textsc{Args}$ for tool-call arguments or
  $\textsc{Result}$ for tool-result content), and the path tokens extracted from that
  channel. View~A is the legacy rule: every path the agent saw counts as surfaced.
  View~B drops only paths whose provenance is a \emph{result} of an engine call
  ($\texttt{codebase\_search}$ or $\texttt{codebase\_graph}$); the engine's
  natural-language query arguments are kept in both views, because the agent did
  produce those tokens. The difference between the two views is the highlighted line.}
  \label{alg:viewb}
  \begin{algorithmic}[1]
    \Require trace $T$ as a sequence of events $(\textit{tid}, \textit{tool}, \textit{role}, \textit{paths})$ where $\textit{role} \in \{\textsc{Args}, \textsc{Result}\}$
    \Ensure $\mathit{surfaced} \subseteq \mathit{files}$
    \Statex \textbf{Engine tools:} $\mathcal{E} \gets \{\texttt{codebase\_search},\, \texttt{codebase\_graph}\}$
    \Statex
    \Function{SurfaceViewA}{$T$}
      \State $S \gets \emptyset$
      \For{$(\textit{tid}, \textit{tool}, \textit{role}, \textit{paths}) \in T$}
        \State $S \gets S \cup \textit{paths}$ \Comment{every surfaced path counts}
      \EndFor
      \State \Return $S$
    \EndFunction
    \Statex
    \Function{SurfaceViewB}{$T$}
      \State $S \gets \emptyset$
      \For{$(\textit{tid}, \textit{tool}, \textit{role}, \textit{paths}) \in T$}
        \If{$\lnot\big(\textit{tool} \in \mathcal{E} \,\land\, \textit{role} = \textsc{Result}\big)$} \Comment{\textbf{the only difference}: engine-result paths skipped; engine args kept}
          \State $S \gets S \cup \textit{paths}$
        \EndIf
      \EndFor
      \State \Return $S$
    \EndFunction
  \end{algorithmic}
\end{algorithm}

We score localization under two views and report under one. \textbf{View~A} (the legacy
rule) counts every path the agent saw as a file the agent reached, including the
candidate paths returned by engine result lists. \textbf{View~B} (agent-targeted)
strips paths whose only provenance is an engine result list, while keeping the engine's
natural-language query arguments. The motivation is that the engine's result list is a
\emph{pointer}, not an arrival: the agent has to choose to grep, read, or edit one of
those candidates for the path to enter the agent's actual trajectory. Treating the
result list as a set of files the agent reached credits SC-ON for being shown a path
while crediting OpenCode for grepping one, an asymmetry that inflates surfaced sets only
for the arm whose engine produces such lists.

Mechanically, the extractor tracks the \texttt{tool\_call\_id} of every surfaced path
and drops the path if and only if its only provenance is a \texttt{codebase\_search} or
\texttt{codebase\_graph} result token. Algorithm~\ref{alg:viewb} states the rule;
View~A and View~B differ in exactly one line. The rule is applied uniformly to all three
arms; SC-OFF and OpenCode emit no engine calls, so the rule is a strict no-op for them
and only SC-ON's numbers move. We adopt View~B as the body's primary view because
stage-decomposed trajectory metrics extend the file-level acc@$k$ posture of LocAgent
\citep{locagent2025} and Agentless \citep{agentless2024} to mixed-action trajectories;
the direction the field has been moving \citep{trajeval2026, sweexplore2026}. The
precise algorithm and the fallback for missing \texttt{tool\_call\_id} are implemented
in the released \texttt{scoring/localization.py}; the full View~A counterpart ships
under \texttt{*\_view\_a} suffixed columns in the released results database.

\FloatBarrier
\subsection{Statistical methods}
\label{sec:design-stats}

Per-instance pass@1 is averaged across the three seeds, giving one value per instance per
arm: for binary outcomes (resolve, acc@$k$) this value lies in $\{0, 1/3, 2/3, 1\}$; for
continuous outcomes (cost, turns, tokens) it is the per-instance mean across seeds.
\textbf{Paired tests} use the \textbf{Wilcoxon signed-rank test} (two-sided, normal
approximation) on those per-instance pass@1 values, between arm pairs. McNemar does not
apply: the per-instance pass@1 metric is no longer binary at the instance level. Per-arm
aggregates are reported as the \textbf{mean of seed means} with the across-seed standard
deviation as a variance estimate. Per-seed values for resolve appear in
Table~\ref{tab:resolve-perseed}. The implementation is pure stdlib and lives in the
released \texttt{analysis/stats.py}; zero differences are dropped under the standard
Wilcoxon convention and no continuity correction is applied (matching
\texttt{scipy.stats.wilcoxon}'s defaults). The body reports \emph{unadjusted}
$p$-values across the ten paired tests in \S\ref{sec:results} (five metrics
$\times$ two arm pairs); readers preferring a family-wise correction at
$\alpha{=}0.05$ can apply Holm or Bonferroni at $m{=}10$ themselves.

\subsection{Pilot disclosure}
\label{sec:design-pilot}

An earlier batch of runs evaluated two additional configurations that did not make it
into the main study. \textbf{Aider} was dropped because it rebuilds the full prompt
(repo-map plus file contents) on every turn, which defeats stable-prefix prompt
caching --- a property structural to Aider's prompt-assembly path rather than a
configuration we could tune around, and one that takes Aider out of the comparable
cost band with the tool-using loops in this study. \textbf{Kimi K2.6} was dropped
because its heavier retrieved context triggered a deterministic stream-decode failure
cluster on heavy instances, taking those cells out of comparability with the rest of
the grid. The pilot cells are released at
\texttt{data/pilot/pilot\_results\_public.\{db,csv\}}.

%% file: tables/tab-metrics.tex
\begin{tabular}{ll}
\toprule
Metric                   & Definition                                                                                  \\
\midrule
resolve                  & $\mathbf{1}[\text{F2P}_{\text{pass}}{=}\text{F2P}_{\text{tot}}{>}0 \land \text{P2P}_{\text{pass}}{=}\text{P2P}_{\text{tot}}]$ \\
$\$/$cell                & per-cell \texttt{cost\_usd} (mean over legit cells, per arm)                                \\
$\$/$solved              & $\sum_{\text{legit}} \text{cost\_usd} \;\big/\; \sum_{\text{legit}} \text{resolved}$, per arm \\
turns                    & tool calls per cell                                                                         \\
tokens                   & LLM input $+$ output tokens per cell                                                        \\
acc@$k$                  & $\mathbf{1}[\,|\text{surfaced}_{1{:}k} \cap \text{gold}| > 0\,]$                             \\
recall@$k$               & $|\text{surfaced}_{1{:}k} \cap \text{gold}| \,/\, |\text{gold}|$                             \\
first-gold rank          & $\min\{\,i : \text{surfaced}_i \in \text{gold}\,\}$ (1-indexed; $\varnothing$ if no match)   \\
\bottomrule
\end{tabular}

%% file: sections/05_integrity.tex
\section{Integrity and Threats to Validity}
\label{sec:integrity}

The integrity load for this study sits in four places: pre-run hardening of the per-cell
sandbox, a post-run audit that re-checked every kept cell for residual leakage, the
exclusion taxonomy that explains which cells were dropped and why, and a named threats list
that points the reader at the residual concerns the audit could not eliminate. The
released artifacts that back this section are \texttt{data/exclusion\_ledger.csv} (the
exclusion ledger), \texttt{scoring/leak\_audit.py} (the auditor), and
\texttt{scoring/localization.py} (the localization extractor that emits both views).

\subsection{Pre-run hardening}
\label{sec:integrity-hardening}

Two mechanisms ran before every cell. \textbf{URL redaction.} Self-links in problem
statements were stripped at spec-build time, so the agent could not navigate to a
canonical-fix page through the prompt. \textbf{Git scrub.} Every cell starts from a
sandbox whose repo has the gold-fix commit and its descendants stripped. Refs deletion
alone is not sufficient: \texttt{git show <hash>} still recovers any object reachable in
the object database, so the hardened path runs \texttt{git gc --prune=now} to physically
remove the dropped objects. A post-scrub object-set check then verifies that no
future-commit objects remain reachable; if any do, the cell is marked
\texttt{scrub=DIRTY} and aborted before the agent runs. Only \texttt{scrub=CLEAN} cells
execute. In the released set, 12 cells tripped this fail-closed gate and appear in the
ledger as \texttt{scrub\_failed}; the trip is whole-instance and arm-independent, so the
distribution is balanced across the three arms (Table~\ref{tab:exclusion-taxonomy}).

\subsection{Post-run audit}
\label{sec:integrity-audit}

After the run completed we re-executed \texttt{leak\_audit.py} over 386 archived
traces from the released set as an S1 reviewer pass. The audit found 5 additional
\texttt{git\_history\_leak} cells: the agent had invoked \texttt{git show <historical-hash>}
on a past commit whose diff touches a gold file. These are commits in the base history
that pre-date the scrub window; the scrub cannot remove them without changing the task
specification. The 5 cells were excluded outcome-blind (any arm, any outcome); the
per-arm breakdown and which were resolved=1 appear in the ledger
(Table~\ref{tab:exclusion-taxonomy}, and \texttt{data/exclusion\_ledger.csv}). Post-exclusion,
the residual impact on every reported metric is under 1\,pp, and no metric ordering flips;
the triple-intersection paired-$n$ moved from 79 to 75. A network-fetch audit was re-run on
the same 386 traces and came back clean (no kept cell fetched a high-severity hosting URL).
The in-ancestry class itself (gold fix reachable in base history) is a dataset limitation
that detect-and-exclude can shrink but not eliminate; the sub-1\,pp figure is what survives
in the released set.

\subsection{Exclusion taxonomy and public ledger}
\label{sec:integrity-ledger}

Twenty-six cells were excluded across the five categories that appear in the public
ledger (Table~\ref{tab:exclusion-taxonomy}). Every excluded cell is removed before
resolve, cost, and localization rates are computed; only legit cells enter denominators
(\S\ref{sec:design-metrics}). Three balance points are worth surfacing.
\texttt{scrub\_failed} is whole-instance balanced across arms, which is the posture the
fail-closed gate is designed to produce. \texttt{provider\_truncation} is arm-asymmetric:
all 5 exclusions land on SuperCoder arms, flagged again in \S\ref{sec:integrity-threats}.
\texttt{git\_history\_leak} (flagged above) and \texttt{leak\_detected} (OpenCode only)
round out the ledger; per-arm counts for every category appear in
Table~\ref{tab:exclusion-taxonomy}. The public artifact
\texttt{data/exclusion\_ledger.csv} carries one row per excluded cell with columns
\texttt{cell\_id}, \texttt{instance\_id}, \texttt{harness}, \texttt{arm},
\texttt{exclusion\_reason}, and \texttt{evidence}; the \texttt{evidence} column quotes
the concrete trigger (for \texttt{git\_history\_leak} rows, the literal
\texttt{git show <hash>} invocation against the gold file). The ledger has been through a
three-check audit pass: structural (every excluded \texttt{cell\_id} resolves in the main
DB), forensic (each \texttt{git\_history\_leak} row hand-confirmed against its trace),
and gold-set verification (every \texttt{instance\_id}'s gold file set matches the
benchmark specification).

\begin{table}[!tbp]
  \centering
  \caption{Exclusion taxonomy. Counts recomputed from
    \texttt{data/exclusion\_ledger.csv}; row totals sum to 26 across 5 categories. All
    excluded cells are removed from rate denominators.}
  \label{tab:exclusion-taxonomy}
  \input{tables/tab-exclusion-taxonomy}
\end{table}

\FloatBarrier
\subsection{Threats to validity}
\label{sec:integrity-threats}

Six residual threats survive the hardening and the audit. \textbf{Paired-$n$ limits.}
Effective denominators are 75 (triple-intersection), 80 (SC-ON vs.\ SC-OFF), and 78
(SC-ON vs.\ OpenCode), defined in \S\ref{sec:design-scope}; resolve-level paired tests
are under-powered at these sizes, and \S\ref{sec:results} states that limitation at every
test. \textbf{Language coverage.} Three of the five SWE-PolyBench languages are
represented (Go, Java, Python); JavaScript and TypeScript were not run, so
generalization beyond these three is not warranted.
\textbf{\texttt{provider\_truncation} asymmetry.} All 5 truncation exclusions fall on the
SuperCoder arms (none on OpenCode); per-arm counts are in
Table~\ref{tab:exclusion-taxonomy}. Because the cells are excluded rather than scored,
the asymmetry does not directly bias the reported rates, but what was lost differs across
arms and is worth flagging. \textbf{Localization extractor sensitivity.} The metric
admits two defensible computations; we report both views throughout
(\S\ref{sec:design-localization}, \S\ref{sec:results}), and the dual-view disclosure is
the mitigation. \textbf{In-ancestry leak residual.} Where the gold fix is reachable in
base history, the scrub cannot remove it without altering the task; detect-and-exclude
(\S\ref{sec:integrity-audit}) shrinks but cannot eliminate this class.
\textbf{Issue-text phrasing realism.} We feed the agent the formal GitHub-issue text from
the upstream benchmarks; \citet{savingswebench2025} show that mutating issue text into
realistic chat-style queries derived from agent telemetry can drop measured pass rates by
over 50\% on some models, an ecological-validity gap that detect-and-exclude does not
address. The reported absolute resolve rates should be read as benchmark-conditional;
the within-harness ablation is robust to this gap because both arms see identical text.

%% file: tables/tab-exclusion-taxonomy.tex
\begin{tabular}{lrrrr}
\toprule
Category                & SC-ON & SC-OFF & OpenCode & Total \\
\midrule
\texttt{scrub\_failed}        & $4$ & $4$ & $4$  & $12$ \\
\texttt{provider\_truncation} & $2$ & $3$ & $0$  & $5$  \\
\texttt{git\_history\_leak}   & $1$ & $2$ & $2$  & $5$  \\
\texttt{leak\_detected}       & $0$ & $0$ & $3$  & $3$  \\
\texttt{install\_failure}     & $0$ & $0$ & $1$  & $1$  \\
\midrule
Total                         & $7$ & $9$ & $10$ & $26$ \\
\bottomrule
\end{tabular}

%% file: sections/06_results.tex
%
%
%

\section{Results}
\label{sec:results}

Three arms (SC-ON, SC-OFF, OpenCode) ran Claude Opus 4.7 across three seeds against the
same SWE-PolyBench Verified and SWE-bench-Pro instances. Each subsection states the point
estimate first and follows immediately with the paired-statistics line. Paired tests use
Wilcoxon signed-rank (two-sided, normal approximation) on per-instance pass@1 across the
three seeds; for binary outcomes (resolve, acc@5) per-instance pass@1 lies in
$\{0, 1/3, 2/3, 1\}$, for continuous outcomes (cost, turns, tokens) it is the
per-instance mean. Per-seed values for resolve appear in
Table~\ref{tab:resolve-perseed}; all other metrics are mean of seed means and appear in
Table~\ref{tab:headline}. Acc@5 is reported under View~B throughout; the methodology and
View~A counterpart are disclosed in \S\ref{sec:results-crossharness}.

\subsection{Cross-harness comparison: SC-ON vs.\ OpenCode}
\label{sec:results-crossharness}

\begin{table}[!tbp]
  \centering
  \caption{Resolve \% per seed (seeds 0, 1, 2) for each arm, with the across-seed mean and
    standard deviation. Direction is consistent across seeds for SC-ON over both
    comparators.}
  \label{tab:resolve-perseed}
  \input{tables/tab-resolve-perseed}
\end{table}

\begin{table}[!tbp]
  \centering
  \caption{Headline metrics across the three arms (mean of seed means across three seeds).
    Acc@5 is View~B (agent-targeted); full tables ship in the released results database.
    Turns, tokens, and wall-clock are per-cell means on legit cells.}
  \label{tab:headline}
  \input{tables/tab-headline}
\end{table}

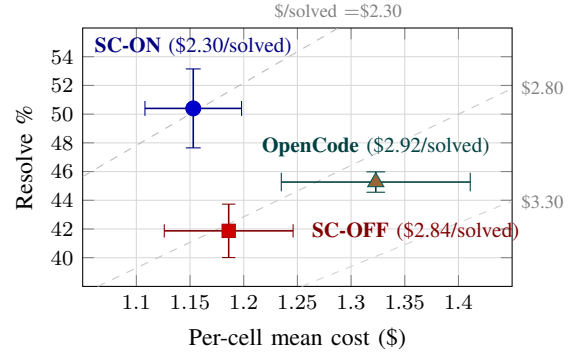
\begin{figure}[!htbp]
  \centering
  \begin{tikzpicture}
    \begin{axis}[
        width=0.82\linewidth, height=5cm,
        xlabel={Per-cell mean cost (\$)},
        ylabel={Resolve \%},
        xmin=1.05, xmax=1.45,
        ymin=38, ymax=56,
        xtick={1.1, 1.15, 1.2, 1.25, 1.3, 1.35, 1.4},
        ytick={40, 42, 44, 46, 48, 50, 52, 54},
        grid=both, grid style={gray!20},
        major grid style={gray!30},
        tick label style={font=\footnotesize},
        label style={font=\small},
        clip=false,
      ]
      \addplot[dashed, gray!55, domain=1.05:1.288, samples=2, forget plot] {100*x/2.30};
      \addplot[dashed, gray!45, domain=1.064:1.45, samples=2, forget plot] {100*x/2.80};
      \addplot[dashed, gray!35, domain=1.254:1.45, samples=2, forget plot] {100*x/3.30};
      \node[gray, font=\scriptsize, anchor=south] at (axis cs:1.288, 56)  {\$/solved $=$\$2.30};
      \node[gray, font=\scriptsize, anchor=west]  at (axis cs:1.45,  51.8) {\$2.80};
      \node[gray, font=\scriptsize, anchor=west]  at (axis cs:1.45,  43.9) {\$3.30};

      \addplot+[only marks, mark=*, mark size=2.8pt, color=blue!70!black,
        error bars/.cd, x dir=both, x explicit, y dir=both, y explicit,
        error bar style={blue!50!black, line width=0.6pt}]
        coordinates {(1.153, 50.40) +- (0.045, 2.75)};
      \node[anchor=south, font=\footnotesize, blue!50!black]
        at (axis cs:1.153, 53.4) {\textbf{SC-ON} (\$2.30/solved)};

      \addplot+[only marks, mark=square*, mark size=2.5pt, color=red!70!black,
        error bars/.cd, x dir=both, x explicit, y dir=both, y explicit,
        error bar style={red!50!black, line width=0.6pt}]
        coordinates {(1.186, 41.87) +- (0.060, 1.86)};
      \node[anchor=west, font=\footnotesize, red!50!black]
        at (axis cs:1.255, 41.87) {\textbf{SC-OFF} (\$2.84/solved)};

      \addplot+[only marks, mark=triangle*, mark size=3.5pt, color=teal!70!black,
        error bars/.cd, x dir=both, x explicit, y dir=both, y explicit,
        error bar style={teal!50!black, line width=0.6pt}]
        coordinates {(1.323, 45.27) +- (0.088, 0.71)};
      \node[anchor=south, font=\footnotesize, teal!50!black]
        at (axis cs:1.323, 46.4) {\textbf{OpenCode} (\$2.92/solved)};

    \end{axis}
  \end{tikzpicture}
  \caption{Cost--resolve plane (mean of seed means; error bars are across-seed standard
  deviation). The three dashed lines are iso-\$/solved reference curves (lines of
  constant cost per solve); steeper means cheaper. SC-ON sits on the \$2.30 curve,
  strictly cheaper than SC-OFF (\$2.84, within-harness) and OpenCode (\$2.92,
  cross-harness). The headline cost claim of the paper is the relative position of the
  SC-ON point, not the per-cell mean alone.}
  \label{fig:pareto}
\end{figure}

Table~\ref{tab:headline} reports the headline metrics for all three arms, and
Figure~\ref{fig:pareto} positions the three arms in the resolve-vs-cost plane: SC-ON
sits upper-left of both comparators, on a cheaper iso-\$/solved curve. Across the three
seeds, SC-ON resolves 50.4\% on average against OpenCode's 45.3\% (paired Wilcoxon on
per-instance pass@1, $n=78$, $\Delta=+6.0$\,pp, $p=0.087$). The direction is consistent
across all three seeds (Table~\ref{tab:resolve-perseed}) but does not separate at the
conventional threshold; the within-harness ablation in \S\ref{sec:results-ablation}
gives the cleanest read.

\textbf{Cost.} SC-ON's \$/solved sits at \$2.30 mean across seeds against OpenCode's
\$2.92 (about 21\% lower). Per-cell mean cost is statistically null (paired Wilcoxon
$p=0.35$); the \$/solved gap is driven by SC-ON's higher resolve rate at comparable
per-cell spend. The substantive claim is that the structural codebase index is not more
expensive to run than agentic grep; on these benchmarks it is favorable.

\textbf{Effort.} SC-ON converges with fewer turns and fewer tokens. Mean turns per cell
are 28.3 for SC-ON against 36.0 for OpenCode (paired Wilcoxon $p<0.0001$); mean tokens
are 10.1k against 14.0k (also $p<0.0001$); mean wall-clock follows at 4.5 against 5.4
minutes. The within-harness ablation in \S\ref{sec:results-ablation} gives the cleanest
mechanism: the index shortens the agent's path to the relevant files, and the saved tool
calls compound into saved tokens and time.

\textbf{Localization.} Acc@5 throughout this paper is View~B (agent-targeted;
\S\ref{sec:design-localization}; the extraction rule is implemented in the released
\texttt{scoring/localization.py}). Under View~B SC-ON acc@5 averages 84.5 across seeds
against OpenCode's 75.3 (paired Wilcoxon $\Delta=+8.1$\,pp, $p=0.080$); under the legacy
View~A the cross-harness comparison shifts (View~A counts engine result-list paths as
files the agent reached, an asymmetry that inflates surfaced sets only for SC-ON). Full
View~A and View~B tables ship in the released results database.

\FloatBarrier
\subsection{Causal ablation: index on vs.\ off}
\label{sec:results-ablation}

\begin{table}[!tbp]
  \centering
  \caption{Causal ablation (SC-ON vs.\ SC-OFF), paired analysis across three seeds. Paired
    tests are Wilcoxon signed-rank (two-sided, normal approximation) on per-instance pass@1
    averaged across seeds. \$/solved is a per-arm aggregate; no paired test applies.
    Acc@5 is View~B.}
  \label{tab:ablation}
  \input{tables/tab-ablation}
\end{table}

Table~\ref{tab:ablation} reports the within-harness ablation. The index is the only thing
that changes between the two arms; model, the rest of the tool set, sandbox, prompts,
seeds, and caps are held fixed.

\textbf{Localization moves substantially.} Under View~B, SC-ON acc@5 averages
\textbf{84.5\%} across seeds against SC-OFF's \textbf{44.3\%} (paired Wilcoxon on
per-instance pass@1, $n=80$, $\Delta=+39.6$\,pp, $p<0.0001$). We use View~B throughout
for the reasons in \S\ref{sec:results-crossharness}. Figure~\ref{fig:firstgold} shows the
discrete CDF behind that point estimate under View~B: the cumulative fraction of legit
cells whose first gold file appears at rank~$\le k$ for $k\in\{1,3,5,10\}$ (the discrete
acc@$k$ values released in the public DB), across all three arms. SC-ON places a gold
file at rank~1 in 77.4\% of cells against SC-OFF's 33.3\%; the gap narrows but does not
close at $k=10$.

\begin{figure}[!tbp]
  \centering
  \begin{tikzpicture}
    \begin{semilogxaxis}[
        width=0.78\linewidth, height=5.5cm,
        xlabel={Rank $k$ (log scale; legit cells with first gold file at rank $\le k$ under View~B)},
        ylabel={Cumulative fraction},
        xmin=0.9, xmax=12,
        ymin=0, ymax=1.0,
        xtick={1, 3, 5, 10},
        xticklabels={1, 3, 5, 10},
        ytick={0, 0.2, 0.4, 0.6, 0.8, 1.0},
        grid=major, grid style={gray!20},
        legend style={at={(0.97, 0.05)}, anchor=south east, font=\footnotesize, draw=gray!50},
        tick label style={font=\footnotesize},
        label style={font=\small},
      ]
      \addplot[blue!70!black, line width=1pt, mark=*, mark size=1.8pt] coordinates {
        (1, 0.774) (3, 0.833) (5, 0.845) (10, 0.905)
      };
      \addlegendentry{SC-ON}
      \addplot[teal!70!black, line width=1pt, dashed, mark=triangle*, mark size=2.2pt] coordinates {
        (1, 0.584) (3, 0.720) (5, 0.753) (10, 0.827)
      };
      \addlegendentry{OpenCode}
      \addplot[red!70!black, line width=1pt, dotted, mark=square*, mark size=1.8pt] coordinates {
        (1, 0.333) (3, 0.398) (5, 0.443) (10, 0.610)
      };
      \addlegendentry{SC-OFF}
      \draw[gray!50, dashed] (axis cs:5,0) -- (axis cs:5,1.0);
      \node[gray, font=\scriptsize, anchor=south] at (axis cs:5, 0.02) {$k{=}5$};
    \end{semilogxaxis}
  \end{tikzpicture}
  \caption{First-gold rank CDF under View~B, per arm. Each marker shows the cumulative
  fraction of legit cells (mean of seed means across three seeds) whose first gold file
  appears at rank~$\le k$ in the agent-targeted surfaced set, for the discrete $k$
  values released in the public DB ($k\in\{1,3,5,10\}$). SC-ON dominates at low ranks
  (the regime that matters for the agent's read budget): 77.4\% of cells place a gold
  file at the top, versus 58.4\% (OpenCode) and 33.3\% (SC-OFF). At $k{=}5$ the values
  equal the body acc@5 (84.5 / 75.3 / 44.3) by construction. For OpenCode and SC-OFF,
  View~A and View~B coincide because neither arm makes context-engine calls.}
  \label{fig:firstgold}
\end{figure}

\textbf{Resolve moves with statistical separation.} SC-ON solves 50.4\% of legit cells
on average across seeds against SC-OFF's 41.9\% (paired Wilcoxon $\Delta=+7.9$\,pp,
$n=80$, $p=0.003$). The direction is consistent across seeds. \S\ref{sec:discussion}
reads the three signals together: large localization gain, statistically separated
resolve gain, no per-cell cost regression.

\textbf{Cost and effort.} The index changes how the agent uses tokens, not how much it
costs per cell. Mean turns drop sharply with the index on (28.3 vs.\ 36.2, paired
Wilcoxon $p<0.0001$); mean tokens drop too (10.1k vs.\ 11.1k, $p=0.027$). Per-cell mean
cost is statistically null (paired Wilcoxon $\Delta=-\$0.118$, $p=0.73$); the index buys
fewer turns without a per-cell cost penalty, and \$/solved comes out lower for SC-ON
(\$2.30 vs.\ \$2.84) because of the higher resolve rate at near-equal per-cell spend.

\FloatBarrier
\subsection{Where the index helps: exploratory heterogeneity}
\label{sec:results-heterogeneity}

\begin{table}[!tbp]
  \centering
  \caption{Resolve and View~B acc@5 by language (Go, Java, Python), mean of seed means
    across three seeds. Exploratory; we make no significance claims.}
  \label{tab:per-language}
  \scriptsize
  \setlength{\tabcolsep}{2pt}
  \input{tables/tab-per-language}
\end{table}

\begin{figure}[!tbp]
  \centering
  \begin{tikzpicture}
    \begin{axis}[
        width=0.78\linewidth, height=5.2cm,
        ybar, bar width=8pt, enlarge x limits=0.30,
        ylabel={View~B acc@5 (\%)},
        symbolic x coords={1-file, 2-file, 3+-file},
        xtick=data,
        ymin=0, ymax=100,
        ytick={0, 20, 40, 60, 80, 100},
        nodes near coords,
        nodes near coords style={font=\scriptsize, /pgf/number format/precision=1, /pgf/number format/fixed, /pgf/number format/fixed zerofill=false},
        legend style={at={(0.5, 1.10)}, anchor=south, legend columns=3, font=\footnotesize, draw=gray!50},
        tick label style={font=\footnotesize},
        label style={font=\small},
        grid=major, grid style={gray!20},
      ]
      \addplot[fill=blue!50!white, draw=blue!70!black] coordinates {(1-file, 85.3) (2-file, 74.1) (3+-file, 91.3)};
      \addlegendentry{SC-ON}
      \addplot[fill=red!40!white, draw=red!70!black] coordinates {(1-file, 42.1) (2-file, 49.0) (3+-file, 44.9)};
      \addlegendentry{SC-OFF}
      \addplot[fill=teal!40!white, draw=teal!70!black] coordinates {(1-file, 74.2) (2-file, 70.4) (3+-file, 81.2)};
      \addlegendentry{OpenCode}
    \end{axis}
  \end{tikzpicture}
  \caption{View~B acc@5 by gold-file count, mean of seed means across three seeds
  ($n{=}46$, $18$, $27$ distinct instances per bucket respectively). The within-harness
  gap (SC-ON minus SC-OFF) is largest in the 3+\,file bucket
  ($46.4$\,pp) where the call-graph intuition predicts that structural ranking pays
  off more than agentic grep. Exploratory; no significance claims.}
  \label{fig:filecount}
\end{figure}
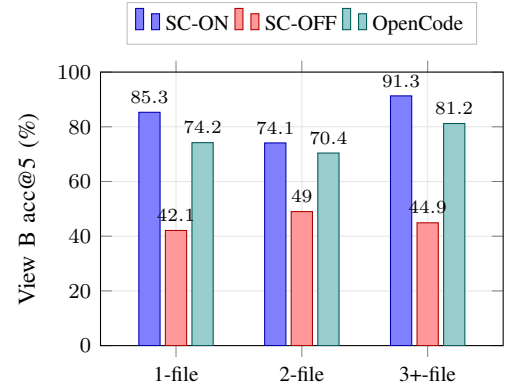

The slices in this subsection are \emph{exploratory}. Per-language bucket $n$ ranges from
18 to 46. We make no significance claims.

\textbf{By language.} Across seeds, localization gains are largest in Go (View~B acc@5
95.4\% ON vs.\ 44.8\% OFF) and Python (82.9\% vs.\ 42.4\%); Java is more modest (71.7\%
vs.\ 46.7\%). Resolve directionally favors SC-ON in every language: Go 47.1\% vs.\
29.9\%, Java 60.0\% vs.\ 53.3\%, Python 47.6\% vs.\ 45.5\%. On Python, the localization
advantage is substantial (82.9\% vs.\ 42.4\%), narrower than the localization gap
suggests but no longer a directional negative on resolve.

\textbf{By gold-file count.} Figure~\ref{fig:filecount} shows acc@5 across the three
arms split by gold-file count. The index's largest gains sit in the 3+\,file bucket
(91.3\% ON vs.\ 44.9\% OFF), consistent with the call-graph intuition: when the change
spans files, a structural ranking pays off more than agentic grep over the working copy.
Per-bucket resolve is directionally positive for SC-ON in every file-count bucket
(1-file 52.7\% vs.\ 41.3\%, 2-file 55.6\% vs.\ 51.0\%, 3+-file 42.0\% vs.\ 36.2\%).

\FloatBarrier
\subsection{Sensitivity}
\label{sec:results-sensitivity}

Full localization tables under both views ship in the released results database.

\textbf{Localization extractor.} Both views are emitted by the released
\texttt{scoring/localization.py} from the same trace set. The DB's canonical \texttt{acc@k}
columns are View~B; View~A is recomputable locally from traces.
The released \texttt{scoring/localization.py} states the extraction rule precisely.

\textbf{Audit residuals.} The 5 outcome-blind \texttt{git\_history\_leak} cells flagged
in the post-run audit (\S\ref{sec:integrity}) move every headline metric by $\le 1$\,pp
when excluded, and no ranking flips. The network-fetch audit was clean.

\textbf{What we do not claim.} The ON vs.\ OpenCode comparison on resolve and View~B
acc@5 is marginal at multi-seed (paired Wilcoxon $p=0.087$ and $p=0.080$ respectively);
the within-harness ablation is the cleaner read. Across-seed standard deviations are
0.7--3.3\,pp on resolve and acc@5 (consistent with recently-reported coding-agent
benchmark noise floors); the reported ablation effects exceed this noise by 3 to
10$\times$.
\FloatBarrier

%% file: tables/tab-resolve-perseed.tex
\begin{tabular}{lccccc}
\toprule
Arm      & Seed 0 & Seed 1 & Seed 2 & Mean & Std \\
\midrule
SC-ON    & 48.8 & 53.6 & 48.8 & \textbf{50.4} & 2.75 \\
SC-OFF   & 43.9 & 41.5 & 40.2 & 41.9          & 1.86 \\
OpenCode & 44.4 & 45.7 & 45.7 & 45.3          & 0.71 \\
\bottomrule
\end{tabular}

%% file: tables/tab-headline.tex
\begin{tabular}{lccc}
\toprule
Metric & SC-ON & SC-OFF & OpenCode \\
\midrule
Resolve \%                          & \textbf{50.4} & 41.9   & 45.3            \\
Loc acc@5 (View B)                  & \textbf{84.5} & 44.3   & 75.3            \\
Recall@5                            & \textbf{0.611}& 0.330  & 0.601           \\
$\$$/solved                         & \textbf{\$2.30} & \$2.84 & \$2.92        \\
$\$$/cell (mean)                    & \textbf{\$1.15} & \$1.19 & \$1.32        \\
Turns (mean)                        & \textbf{28.3} & 36.2   & 36.0            \\
Tokens, k (mean)                    & \textbf{10.1} & 11.1   & 14.0            \\
Wall-clock, min (mean)              & \textbf{4.5}  & 5.5    & 5.4             \\
\bottomrule
\end{tabular}

%% file: tables/tab-ablation.tex
\begin{tabular}{lcccc}
\toprule
Metric                              & SC-ON          & SC-OFF & $\Delta$                       & Paired $p$ \\
\midrule
Resolve \%                          & 50.4           & 41.9   & $\mathbf{+7.9}$\,\textbf{pp}   & $\mathbf{0.003}$    \\
Loc acc@5 (View B)                  & \textbf{84.5}  & 44.3   & $\mathbf{+39.6}$\,\textbf{pp}  & $\mathbf{<0.0001}$  \\
Recall@5                            & \textbf{0.611} & 0.330  & $\mathbf{+0.281}$              & $\mathbf{<0.0001}$  \\
Turns (mean)                        & 28.3           & 36.2   & $-8.3$                         & $\mathbf{<0.0001}$  \\
Tokens, k (mean)                    & 10.1           & 11.1   & $-1.6$                         & $0.027$             \\
$\$$/cell (mean)                    & \$1.15         & \$1.19 & $-\$0.118$                     & $0.73$ (null)       \\
$\$$/solved                         & \$2.30         & \$2.84 & $-\$0.54$                      & n/a                 \\
\bottomrule
\end{tabular}

%% file: tables/tab-per-language.tex
\begin{tabular}{lcccccc}
\toprule
          & \multicolumn{3}{c}{Resolve \%} & \multicolumn{3}{c}{Loc acc@5 (View B)} \\
\cmidrule(lr){2-4} \cmidrule(lr){5-7}
Language ($n_{\text{ON/OFF/OC}}$)  & SC-ON & SC-OFF & OpenCode & SC-ON & SC-OFF & OpenCode \\
\midrule
Go (29/29/31)     & \textbf{47.1} & 29.9 & 35.5 & \textbf{95.4} & 44.8 & 86.0 \\
Java (20/20/19)   & \textbf{60.0} & 53.3 & 57.9 & \textbf{71.7} & 46.7 & 66.7 \\
Python (35/33/31) & \textbf{47.6} & 45.5 & 47.3 & \textbf{82.9} & 42.4 & 69.9 \\
\bottomrule
\end{tabular}

%% file: sections/07_discussion.tex
%
%

\section{Discussion}
\label{sec:discussion}

\textbf{What the ablation says.} SC-ON and SC-OFF differ only in the two engine tools
(\S\ref{sec:design-arms}, \S\ref{sec:system-engine}); model, prompts, sandbox, scorer,
seeds, caps, and the remaining tool set are held identical, so the within-harness deltas
read causally on the index. View~B acc@5 moves from 44.3\% to 84.5\% (paired Wilcoxon
$p<0.0001$), and resolve moves from 41.9\% to 50.4\% (paired $p=0.003$); per-cell mean
cost is statistically null ($p=0.73$) while turns and tokens both fall significantly. The
substantive read is that the index moves localization a lot, moves resolve significantly,
and is cost-neutral at the cell level while favorable on \$/solved.

\textbf{Cross-harness validity.} SC-ON matches or modestly favors OpenCode on resolve
($p=0.087$) and on View~B acc@5 ($p=0.080$), directionally in SC-ON's favor across seeds
but not at conventional significance. The structural codebase index does not duplicate
behavior that competent agentic grep already reaches; at minimum, against a competent
grep-and-read comparator, the index does not regress the agent on the metrics that decide
the task. The cross-harness effort gap is where the two arms separate cleanly: SC-ON
converges in fewer turns and fewer tokens (both $p<0.0001$, magnitudes in
Table~\ref{tab:headline}).

\textbf{Where the index does not help.} The ON vs.\ OpenCode comparisons on resolve and
View~B acc@5 are marginal at conventional significance thresholds (\S\ref{sec:results-crossharness});
the within-harness ablation is the cleanest read of what the index actually contributes.

\textbf{Implications for harness design.} We treat the implications as conditional
observations, not prescriptions. The largest causal localization gain in our data sits in
the 3-or-more-file gold bucket (\S\ref{sec:results-heterogeneity},
Figure~\ref{fig:filecount}), consistent with the call-graph intuition: when the change
spans files, a structural index that ranks paths by reachability pays off more than
agentic grep over the working copy. The cost-favorable finding simplifies the deployment
question: at a fixed model and comparable caps, a structural codebase index is the lower
\$/solved arm on these benchmarks, so the per-task win comes from localization quality
and the downstream turn savings it enables.

\paragraph{Conclusion.} This study reports a leak-audited, model-controlled, causal
ablation of a shipped structural codebase index inside a coding-agent harness, paired
with a cross-harness validity check against an agentic-grep comparator. The index
causally moves localization substantially and resolve with statistical separation
(50.4\% vs.\ 41.9\% on resolve, paired Wilcoxon $p=0.003$), at no per-cell cost penalty
and lower \$/solved than either the within-harness OFF arm or the OpenCode comparator. The released artifacts (the
exclusion ledger, the audit script, and the dual-view localization extractor) back the
integrity claims in \S\ref{sec:integrity} and the result numbers in \S\ref{sec:results}.
At this model and on these benchmarks, the deployment question for a structural codebase
index is not whether it is too expensive to run, but whether the workload includes
multi-file changes where structural ranking pays off.